\documentclass[conference]{IEEEtran}

\IEEEoverridecommandlockouts
\usepackage{cite}
\usepackage{amsmath,amssymb,amsfonts}
\usepackage{algorithmic}
\usepackage{graphicx}
\usepackage{textcomp}
\usepackage{subfig} 
\usepackage{xcolor}

\usepackage{xcolor}

\usepackage{mwe}

\def\BibTeX{{\rm B\kern-.05em{\sc i\kern-.025em b}\kern-.08em
    T\kern-.1667em\lower.7ex\hbox{E}\kern-.125emX}}
\begin{document}

\title{Closer Look at the Uncertainty Estimation in Semantic Segmentation under Distributional Shift\\
\thanks{This work has been partially supported by Statutory Funds of Electronics, Telecommunications and Informatics Faculty, Gdańsk University of Technology. This work was supported in part by the Polish National Centre for Research and Development (NCBR) through the European Regional Development Fund entitled: INFOLIGHT Cloud-Based Lighting System for Smart Cities under Grant POIR.04.01.04/2019.}
}


\author{\IEEEauthorblockN{Sebastian Cygert\textsuperscript{\textsection}}
\IEEEauthorblockA{\textit{Multimedia Systems Department} \\
\textit{Gdańsk University of Technology}\\
Gdańsk, Poland \\
sebcyg@multimed.org}\\
\and
\IEEEauthorblockN{Bartłomiej Wróblewski\textsuperscript{\textsection}}
\IEEEauthorblockA{
\textit{Gdańsk University of Technology}\\
bart.wroblew@gmail.com}\\
\IEEEauthorblockN{Karol Woźniak}
\IEEEauthorblockA{
\textit{Gdańsk University of Technology}\\} \\
\and
\IEEEauthorblockN{Radosław Słowiński}
\IEEEauthorblockA{
\textit{Gdańsk University of Technology}\\} \\
\IEEEauthorblockN{Andrzej Czyżewski}
\IEEEauthorblockA{\textit{Multimedia Systems Department} \\
\textit{Gdańsk University of Technology}\\} \\

}

\maketitle

\maketitle
\begingroup\renewcommand\thefootnote{\textsection}
\footnotetext{Equal contribution}
\endgroup

\begin{abstract}
While recent computer vision algorithms achieve impressive performance on many benchmarks, they lack robustness - presented with an image from a different distribution, (e.g. weather or lighting conditions not considered during training), they may produce an erroneous prediction. 
Therefore, it is desired that such a model will be able to reliably predict its confidence measure.
In this work, uncertainty estimation for the task of semantic segmentation is evaluated under a varying level of domain shift: in a cross-dataset setting and when adapting a model trained on data from the simulation. It was shown that simple color transformations already provide a strong baseline, comparable to using more sophisticated style-transfer data augmentation. Further, by constructing an ensemble consisting of models using different backbones and/or augmentation methods, it was possible to improve significantly model performance in terms of overall accuracy and uncertainty estimation under the domain shift setting.  The Expected Calibration Error (ECE) on challenging GTA to Cityscapes adaptation was reduced from 4.05 to the competitive value of 1.1.  Further, an ensemble of models was utilized in the self-training setting to improve the pseudo-labels generation, which resulted in a significant gain in the final model accuracy, compared to the standard fine-tuning (without ensemble).

\end{abstract}

\begin{IEEEkeywords}
uncertainty estimation, semantic segmentation, domain adaptation, self-training, ensemble of models
\end{IEEEkeywords}

\section{Introduction}

In recent years, visual recognition has witnessed impressive progress on many benchmarks. However, the application of deep learning methods for agents operating in the real world, e.g. autonomous driving, is still limited. A significant challenge is that current vision models lack robustness \cite{concrete}. It has been shown that current CNN-based models are sensitive to a novel type of noise \cite{dodge}, changes in context\cite{objectnet}, temporal changes in video \cite{temporal} and novel weather conditions \cite{winter}. These examples show that CNNs are sensitive to distributional shift: when the test-time distribution of the data differs from the training distribution. Additionally, current models seem to be biased towards texture information \cite{cnnbiased}, largely ignoring shape information. This again can be very dangerous for real-world deployment, for example in the case of sensor noise \cite{HendrycksD19}. 

What is more, current models tend to be overconfident in their outputs \cite{calibration}. The problem is even more evident for the distributional shift\cite{trustuncertainty}. For models operating in the real world, it is of great importance to be robust to such distributional changes, because for many applications it is not possible to collect a large and diverse dataset that contains all possible situations that may occur during deployment (e.g., new weather or lighting conditions, different types of distortions).

A task of particular importance for agents operating in the real world is reliable uncertainty estimation, which can be beneficial in many ways. During deployment, the agent could warn that its prediction is not reliable (medicine), or could effectively integrate predictions from different modalities (autonomous driving) \cite{multi}. Uncertainty estimation could also be used for pseudo-labelling of unlabelled data, to further improve model accuracy in the target domain in a self-training setting \cite{noisystudent}.

This work focuses on studying uncertainty estimation for semantic segmentation, which is a very important task with significant application potential. Further, our study focuses on distributional shift, which is essential importance for real-world deployment. We study uncertainty calibration in different settings:
\begin{itemize}
    \item when a model trained on the simulation is tested on real-world data (large distributional shift)
    \item cross-dataset evaluation (mild distributional shift)
\end{itemize}

Further, we utilized a state-of-the-art method for model calibration, namely an ensemble of models \cite{trustuncertainty, ensemblesuncertainty}, to improve the model calibration. This allows the calibration of predictive uncertainty (and overall accuracy) to be significantly improved, especially under domain shift. Finally, we show the effect of using an ensemble of models on downstream task of domain adaptation, for which we utilize a popular self-training approach \cite{noisystudent,textureinvariant,selfeccv2018}. Our study is aimed at the reality-check for uncertainty estimation and domain adaptation methods. Studying the performance for the varying domain shift for the aforementioned methods is vital empirical work for real-world applications. We have focused on autonomous driving due to the availability of large annotated datasets from both simulations and the real world, and potential applications. Our contributions are as follows:
\begin{itemize}
    \item We study how the model accuracy and uncertainty estimation for semantic segmentation are affected by the varying level of distributional shift. Further, an ensemble of models approach is evaluated in the same setting.
    \item We show that simple color transformations can be as effective as style-transfer data augmentation for increasing models’ robustness.
    \item We show how an ensemble of models can be utilized in the self-training approach to improve model adaptation to the target domain further.
\end{itemize}

\section{Related work}

\textbf{Robustness.} Evaluating models in an out-of-distribution setting, where the test-time dataset is from a different distribution than the training data, is essential for real-world applications \cite{trustuncertainty,concrete, semanticbenchmark}. This is because machine learning models might provide incorrect predictions when presented with, for example, noisy data, or different lighting or weather conditions \cite{HendrycksD19}. Several methods based on data augmentation have been proposed to improve models' robustness in visual recognition, from which style-transfer data augmentation is very popular \cite{cnnbiased, Cygert}. In our work, style-transfer data augmentation was utilized, but we also noticed that simply applying color-jittering during training can be beneficial for the cross-dataset evaluation, which confirms the recent finding that very simple naturalistic augmentation can be very effective \cite{todorealistic, robcompr}.

\textbf{Uncertainty estimation.}
One of the problems with modern neural networks is that they are poorly calibrated and tend to be overconfident in the predictions \cite{calibration}. Different techniques exist for improving estimates of predictive uncertainty. A classical approach is called temperature scaling, where the model confidences are scaled using a post-hoc procedure on the held-out validation set \cite{temperature}. A popular approximate Bayesian approach is a dropout-based model, where the predictive uncertainty is computed based on the multiple outputs of the model on a given image (with dropout enabled) \cite{Gal}. Another sampling-based approach uses agreement between an ensemble of models as a measure of model uncertainty \cite{ensemble}. Interestingly, using ensembles has been shown to yield the best results on uncertainty estimation under the distributional shift \cite{trustuncertainty, ensemblesuncertainty}. 
The ensembles' common setup is to use neural networks trained using different random initialization weights to induce diversity between the models \cite{ensembles_1990}. This is because it has been shown that networks pre-trained on the same dataset stay in the same basin in the loss landscape, and thus reduce variation in the models \cite{TODOnipstransfer}. However, we found that semantic segmentation models trained on used datasets using random initialization perform rather poorly. As such, we show that it is possible to create an efficient model ensemble using models with different backbones and data augmentations.

\textbf{Domain adaptation.}
While it is a standard to evaluate machine learning models on i.i.d. (independent and identically distributed) data, for the real-world deployment, the data may come from different distribution than the training data. As such, many methods for domain adaptation have been proposed which use unlabelled data from the target domain to improve the accuracy of the model. Popular approaches include matching image statistics between domains \cite{histogram}, learning shape-based representation \cite{textureinvariant}, self-learning \cite{textureinvariant} and using data from simulation \cite{james2019simtoreal, synthetic}. 
Using simulated data in particular is interesting, since a simulation allows numerous and diverse training examples to be generated. Simultaneously, the difference in data distribution between the source and target domains is very challenging for real-world problems, and sometimes using labelled data can actually hinder performance \cite{todorealistic}. As such, it is important to evaluate the models’ performance under varying levels of distributional shift.

\textbf{Self-training.} In our work, we make use of the self-learning method, which works in two stages. First, given a trained model, confident predictions are gathered for the target domain, which are also called pseudo-labels. In the next stage, the pseudo-labels are used to fine-tune the model, which allows for domain adaptation. The potential problem with self-learning is that the gathered pseudo-labels might contain erroneous predictions. As such, we propose to use an ensemble approach to gather the pseudo-labels, as ensembles are known to have both good accuracy and uncertainty estimation, which are crucial for the efficient pseudo-labelling stage.

Similar to our work, \cite{medicalcaluncertainty} shows that an ensemble of models is efficient for improving uncertainty estimation in medical image segmentation. We additionally show the effect of ensembles under distributional shift and their utility for the downstream task of domain adaptation. Ensemble predictions on unlabelled datasets were also used as soft targets for direct training supervision for the classification problem \cite{modelcompression}. Here, we use an alternative approach with hard labels, where the least confident predictions are discarded during training, for the task of semantic segmentation. 
 
A similar approach to ours \cite{textureinvariant} uses style-transfer data augmentation to train a base model, which is further adapted to the target domain using self-training. We show that simpler data augmentations can also be very efficient, and that ensemble of models makes the fine tuning stage more efficient which, in turn, makes our work complementary. 

\section{Methodology}
\subsection{Semantic Segmentation}
Semantic segmentation can be viewed as a pixel-wise classification problem where the goal is to assign to each pixel a predicted category $c \in \{1, ..., C\}$. As it is now common in the visual recognition area, semantic segmentation models are mostly based on Convolutional Neural Networks (CNNs), for example, FCN \cite{FCN}. As it is a classification problem, standard cross-entropy loss can be used to optimize the model weights over the training images:
\begin{equation}
    L_{CE} = -\frac{1}{N} \sum_{i=1}^{N} \sum_{c=1}^{C} (y_i = c) log (p(\hat{y_i} = c))
\end{equation}
where $(y_i = c) \in \{0,1\}$ indicates whether class $c$ is the correct class for pixel $i$ and $\hat{y_c}$ is a predicted probability for class $c$ at pixel $i$, and $N$ is the number of pixels. For each pixel model returns a logit vector $ z_i \in R_c$. Further a $softmax$ function is applied $p_i = softmax(z_i)$, which returns a list of predicted class probabilities for a given pixel. The class with the highest probability is used as the predicted class with an associated probability score.

Over the years, many different architectures have been developed, and in our work, we used DeepLabV3+ \cite{deeplab}, which is commonly used in the community. Furthermore, different backbones (e.g. the large ResNet-101 or the lightweight MobileNet) can be used according to the requirements. 

For evaluation, two metrics were used. Pixel accuracy simply measures what percentage of pixels are correctly predicted. Another popular metric is the mean IoU (intersection over union). The IoU metric is computed for each class and then the mean value (mIoU) is reported.

\subsection{Uncertainty Estimation}
An output of the semantic segmentation network is a predicted class $c$ for each pixel, associated with confidence value $p$. Ideally, such classifier would be well-calibrated, thus correct predictions should be associated with high confidence and poor predictions contrary. One of the ways to measure model calibration is to compute an Expected Calibration Error (ECE) \cite{calibration}. To compute the ECE score, pixel-wise predictions are partitioned into $m$ equally-sized bins based on the confidence value, and the ECE score is computed as the difference between the average confidence and the average accuracy in each bin, weighted by the number of predictions in each bin:
\begin{equation}
ECE = \sum_{m=1}^M \frac{|B_m|}{n}|acc(B_m) - conf(B_m)|
\end{equation}
where $B_m$ is the set of indices that falls into the \textit{m-th} bin. Intuitively, when a well-calibrated segmentation network outputs a 90\% confidence value for some set of pixels, it should be correct in 90\% of the cases. The lower the value, the better the calibration that is obtained (0 means perfect calibration). 

\subsection{Ensemble of models}
To improve model calibration, we utilized the model ensemble method which has been shown to provide the best results among other methods, especially under distributional shift \cite{trustuncertainty}. In the case of the ensemble approach, it is common to train models using randomly initialized networks to induce diversity between models \cite{ensembles_1990}. However, we found that semantic segmentation models trained on the GTA or Cityscapes dataset without pretraining performs rather poorly. As a result, we used ImageNet pretraining. However, to achieve diversity between the models, different backbones and/or augmentation methods were used. It was shown in the literature that using as few as 5 models can provide very good results \cite{trustuncertainty}, and because of the computational budget, we used 5 models in our experiments. Namely, given $M$ independently trained models, a final semantic segmentation $p_E$ for the image $x$ can be computed as the average of all models predictions:

\begin{equation}
p_E(x) = \frac{1}{M} \sum_{i=1}^M p_m(x)
\end{equation}
where $p_m$ is the prediction of the \textit{m-th} model in the ensemble.

\subsection{Data augmentation}
To improve the models’ adaptation to the distributional shift, a style-transfer data augmentation was utilized, which has been shown to improve model robustness \cite{cnnbiased, Cygert}. As the source of style images, Kaggle's \textit{Painter By Numbers}\footnote{https://www.kaggle.com/c/painter-by-numbers/} dataset was used, similar as in \cite{cnnbiased}, and during training a stylized image was sampled with probability $p = 0.5$, otherwise, the original image was used. To generate stylized datasets, we have used the method presented in the literature \cite{style_transfer}. 

We also hypothesized that using simple color transformations could also be beneficial in the domain adaptation setting as it would make the model more invariant to the texture information. As, as an alternative to the style-transfer, the following color jittering transformations from the Tensorflow API\footnote{https://www.tensorflow.org/api\_docs/python/tf/image} were also used during training: random changes in the brightness, contrast, saturation, and hue of the images. Details are described in the implementation details section.

\begin{figure}[t]
\centering
\subfloat{{\includegraphics[width=0.5\linewidth]{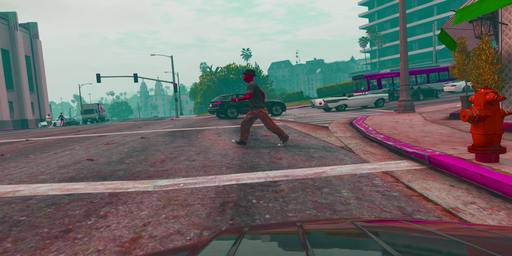} }} 
\subfloat{{\includegraphics[width=0.5\linewidth]{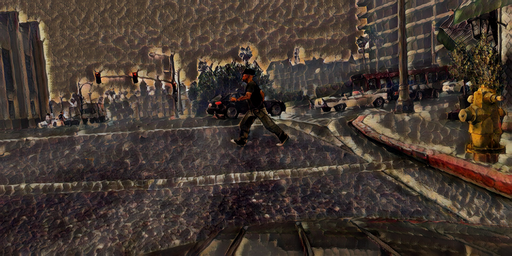} }}
\vspace{-3mm}
\subfloat{{\includegraphics[width=0.5\linewidth]{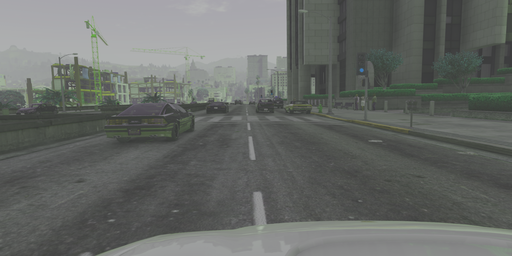} }} 
\subfloat{{\includegraphics[width=0.5\linewidth]{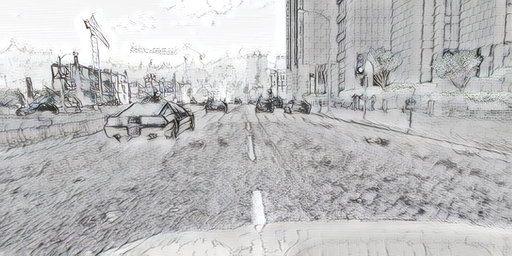} }}
\hspace{0mm}
\caption{Different augmentation strategies applied to sample images from the GTA dataset. First column - color transformations, second column - style transfer.}
\label{fig:fig1}
\end{figure}

Using different augmentation strategies could also be beneficial in the ensemble, as the models trained with different augmentations might learn different representations. Fig. \ref{fig:fig1} shows examples of augmented images.

\section{Experiments}

\subsection{Datasets}
For our experiments, popular semantic segmentation datasets were used. They contain dense pixel-level semantic annotations for the same 20 classes (including the “ignore” class – usually representing the background). 

\textbf{GTA}\cite{playingfordata} is a dataset for which data were collected in a simulated environment, i.e. a modern computer game. It consists of 22,466 training and 2500 validation images, and is commonly used to evaluate simulation-to-real transfer. \textbf{Cityscapes} \cite{cityscapes} is a popular autonomous driving dataset for which data was collected in 27 cities in Germany, consisting of 2975 training images and 500 validation images. Although Cityscapes is a diverse dataset, a potential limitation is the fact that the data was collected mostly during the daytime in good weather conditions. The \textbf{Berkeley Deep Drive (BDD)} dataset \cite{bdd} provides data collected in diverse weather conditions (e.g. rain, snow), scene types (city, highway, countryside), and also images recorded during nigh-time. Pixel-level annotations are provided for 10,000 training and 1000 validation images. 

In our experiments, we focussed on domain adaptation from simulation to real data (GTA-to-Cityscapes) and cross-dataset evaluation (Cityscapes-to-BDD). 

\subsection{Implementation details}
For all experiments, the DeepLabv3+\cite{deeplab} network was used with different backbones (ResNet-101, Xception41, Xception65) pre-trained on ImageNet. Specifically, all models were trained on 2 GPUs for 100,000 steps with a batch size of 16.  As in the original paper, a polynomial decay learning rate was used with an initial learning rate = 0.01 and parameter \textit{power} set to 0.9. 

The data augmentations are consistent with the official implementation\footnote{https://github.com/tensorflow/models/tree/master/research/deeplab}, specifically random scaling (in the range 0.5 to 2.0) and left-right flipping were applied during the training procedure. All images were rescaled to the size 512 x 1024 pixels. Color jittering was applied using TensorflowAPI with the following transformations: random brightness (adjustment factor in the range [0, 0.25)), random contrast (contrast factor in the range [0.5, 1.5)), random saturation (saturation factor in the range [1.0, 3.0)), and random hue (hue offset in the range [0, 0.25)). The chosen hyperparameters were experimentally validated to provide visually diverse images.

All models were evaluated using the validation sets (as the test sets’ ground-truth data are not publicly available). During the fine-tuning stage, the models were trained for 25,000 steps as we noticed that the training loss converged around 20,000 steps for all of the models. When reporting the results, CJ stands for a model trained using color jittering transformations, while SIN stands for a model trained using style-transfer, as in \cite{cnnbiased}.

\subsection{Baseline models}

First, the DeepLabV3+ model with the ResNet backbone was trained on both the GTA and Cityscapes datasets and further evaluated (see Table \ref{tab:tab1}). Several observations can be made. First, there was a very significant drop in accuracy when the models were evaluated under domain shift. The gap was larger for sim-to-real adaptation (GTA-to-Cityscapes) compared to the cross-dataset evaluation (Cityscapes-to-BDD). Further, it can be observed that evaluated data augmentations only slightly affected the performance on the source domain, but they showed really impressive performance in the domain adaptation setting. For GTA-to-Cityscapes, the mIoU increased from 25.4 to 40.4, and similarly for Cityscapes-to-BDD, the mIoU rose from 42.8 to 49.1. Nevertheless, the domain gap was still quite large; a model trained on the Cityscapes dataset achieved a mIoU of 74.1, compared to 40.4 achieved by a model trained on the GTA dataset.

Noticeably, applying simple color transformations worked as well as using an advanced technique of style-transfer, which is consistent with a very recent finding \cite{origins}. Looking at the model calibration, one can notice that all of the models were almost perfectly calibrated when evaluated on the source domain, however when evaluated under domain shift, the ECE metric greatly increased, e.g. for a model trained on the Cityscapes dataset, the metrics increased from 1.49 to 9.55 when evaluated on the BDD dataset instead of Cityscapes. Consistent with recent findings, it was shown that using texture-based data augmentation improved model calibration under domain shift \cite{Cygert}, with the SIN model obtaining slightly better results than using color transformations. In general, using any of aforementioned data augmentations was crucial in the domain adaptation setting.

\begin{table}[t]
\centering
\caption{Performance of DeepLabv3 using ResNet-101 backbone under different evaluation settings. CJ models were trained using color jittering and SIN models used style-transfer augmentation.
}
\label{tab:tab1}
\begin{tabular}{|l|l|l|l|l|l|l|}
\hline
Model name & mIoU & pix. acc & ECE & mIoU & pix. acc & ECE\\
\hline
 & \multicolumn{3}{|c|}{\textbf{GTA-to-GTA}} & \multicolumn{3}{|c|} {\textbf{GTA-to-Cityscapes}} \\
\hline
Baseline & \textbf{80,8} & \textbf{96,6} & \textbf{0.16} & 25,4 & 60,1 & 23.36 \\
\hline
CJ & 80,6 & 96,4 & 0.21 & \textbf{40,4} & 83,7 & 6.5\\
\hline
SIN & 77,2 & 95,9 & 0.21 & 40 & \textbf{83,9} & \textbf{5.06}\\
\hline
\hline
 & \multicolumn{3}{|c|}{\textbf{Cityscapes-to-Cityscapes}} & \multicolumn{3}{|c|}{\textbf{Cityscapes-to-BDD}} \\
\hline
\hline
Baseline & 74,1 & \textbf{95,5} & 1.49 & 42,8 & 83,9 & 9.55 \\
\hline
CJ & \textbf{74,4} & 95,4 & 1.36 & 49,1 & 89,4 & 5.07 \\
\hline
SIN & 71,4 & 95 & \textbf{1.18} & \textbf{49,3} & \textbf{89,8} & \textbf{4.56}\\
\hline
\end{tabular}
\end{table}

\subsection{Model calibration}

An ensemble of models method was used to improve model calibration, which utilized three different backbones (ResNet-101, Xception41, Xception65) and two different augmentation methods (color jitter and style-transfer). We also experimented with the PNAS architecture \cite{pnas}, which is known to achieve great accuracy, however, the performance was not satisfactory, as no pre-trained model is currently available for that model.  Table \ref{tab:models} shows the performance in the cross-dataset setting for the Xception models. Comparing this to Table \ref{tab:tab1}, one can see that Xception models performed slightly better than models using ResNet-101 as the backbone.

\begin{table}[t]
\centering
\caption{Xception models performance under cross-dataset setting.
}
\label{tab:models}
\begin{tabular}{|l|l|l|l|}
\hline
Name & mIOU & pix. acc & ECE\\
\hline
 \multicolumn{4}{|c|}{GTA-to-Cityscapes} \\
\hline
Xception41 (CJ) & 41.8 & 82.7 & 7.3 \\
\hline
Xception41 (SIN) & \textbf{43.7} & \textbf{86.2} & \textbf{4.05} \\
\hline
Xception65 (CJ) & 41.3 & 82.0 & 7.47 \\
\hline
\hline
\multicolumn{4}{|c|}{Cityscapes-to-BDD} \\
\hline
\hline
Xception41 (CJ) & \textbf{52.6} & 90.3 & 4.5\\
\hline
Xception41 (SIN) & 51.1 & \textbf{90.9} & \textbf{3.74} \\
\hline
Xception65 (CJ) & 52.4 & 90.4 & 5.09 \\
\hline
\end{tabular}
\end{table}

Table \ref{tab:ensembles} shows ensemble performance. 
Our base ensemble consists of $M=3$ models, for which color jittering transformations were used during training using ResNet-101 and Xception backbones. Further, two additional models were trained using style transfer augmentation (ResNet-101 and Xception41 backbones) and results in another ensemble variant with $M=5$ models. While the results with no domain shift were comparable, obtained results were better under the domain shift when using 5 models. The mIoU increased from 43.2 to 44.5, and from 55.7 to 56.2 on the Cityscapes and BDD datasets, respectively. Similarly, the ECE was significantly reduced on both datasets. It is also very important to notice that the ensemble performance was better than its strongest member, i.e. for the Cityscapes-to-BDD transfer, the strongest single model obtained a mIoU of 52.6 (Xception41 - CJ), while the ensemble accuracy was 56.2. Similarly, the ECE significantly improved for the ensemble under domain shift: for the GTA-to-Cityscapes transfer, the ensemble ECE was 1.09, while the best result from a single model was 4.05 (Xception41 – SIN).
Additionally, Fig. \ref{fig:calib} shows the calibration plot, comparing the model calibration of our highest-capacity model (Xception-65) with the calibration of the ensemble. Overall, it was confirmed that our ensemble improved both accuracy and uncertainty calibration, especially under domain shift.

\begin{table}[t]
\centering
\caption{Ensemble of models performance. Also mean performance of all models is reported.
}
\label{tab:ensembles}
\begin{tabular}{|l|l|l|l|l|l|l|}
\hline
Model name & mIoU & pix. acc & ECE & mIoU & pix. acc & ECE\\
\hline
 & \multicolumn{3}{|c|}{\textbf{GTA-to-GTA}} & \multicolumn{3}{|c|} {\textbf{GTA-to-Cityscapes}} \\
\hline

M=3 & \textbf{81.9} & \textbf{96.8} & 0.81 & 43.2 & 84.7 & 2.45 \\
\hline
M=5 & 81.4 & 96.7 & 1.02 & \textbf{44.5} & \textbf{86.3} & \textbf{1.1}\\
\hline
Models mean & 79.0 & 96.3 & \textbf{0.21} & 41.4 & 83.7 & 6.08 \\
\hline
\hline
 & \multicolumn{3}{|c|}{\textbf{Cityscapes-to-Cityscapes}} & \multicolumn{3}{|c|}{\textbf{Cityscapes-to-BDD}} \\
\hline
\hline
M=3 & \textbf{77.2} & \textbf{96.0} & 0.36 & 55.7 & 91.3 & 1.99 \\
\hline
M=5 & 77.0 & \textbf{96.0} & \textbf{0.29} & \textbf{56.2} & \textbf{91.7} & \textbf{1.09} \\
\hline
Models mean & 73.8 & 95.4 & 1.16 & 49.3 & 89.8 & 4.56 \\ 
\hline
\end{tabular}
\end{table}

\newlength{\tempheight}
\newlength{\tempwidth}

\newcommand{\rowname}[1]
{\rotatebox{90}{\makebox[\tempheight][c]{\textbf{#1}}}}
\newcommand{\columnname}[1]
{\makebox[\tempwidth][c]{\textbf{#1}}}

\begin{figure}[ht!]
\setlength{\tempwidth}{.47\linewidth}
\settoheight{\tempheight}{\includegraphics[width=\tempwidth]{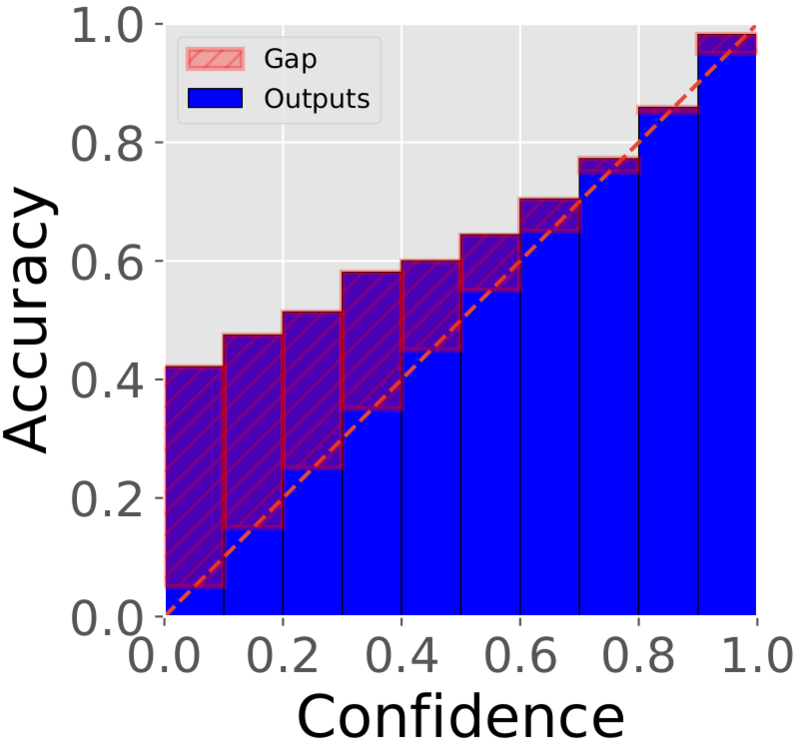}}%
\centering
\columnname{Xception65}\hfil
\columnname{Ensemble}\\
\subfloat{\includegraphics[width=\tempwidth]{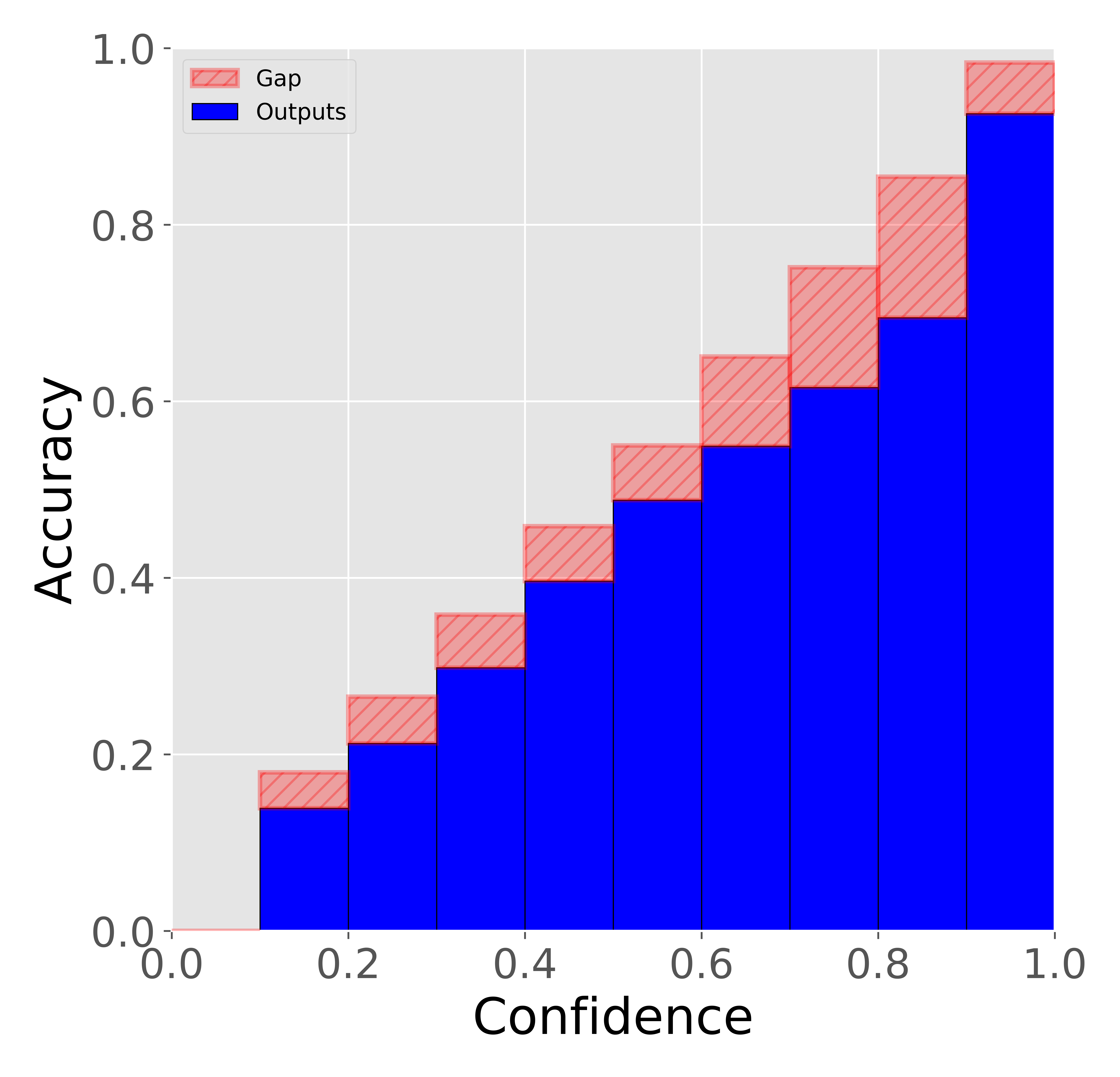}}\hfil
\subfloat{\includegraphics[width=\tempwidth]{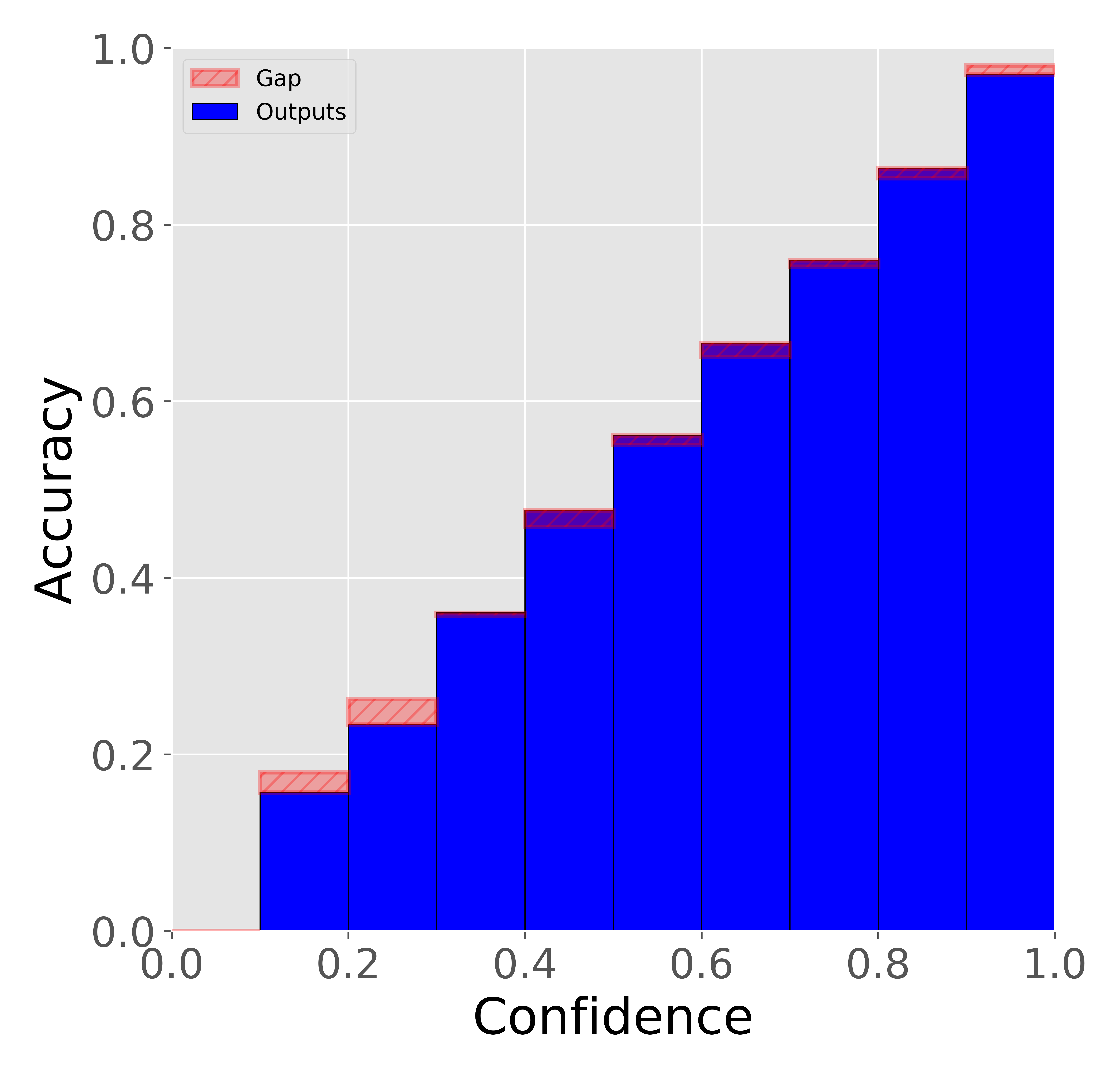}}\hfil
\caption{Calibration plots for Xception65 model and model ensemble (M=5) evaluated on the GTA-to-Cityscapes adaptation. Note great calibration for the ensemble of models.}
\label{fig:calib}

\vspace*{\floatsep}%

    \centering
    \includegraphics[width=\linewidth]{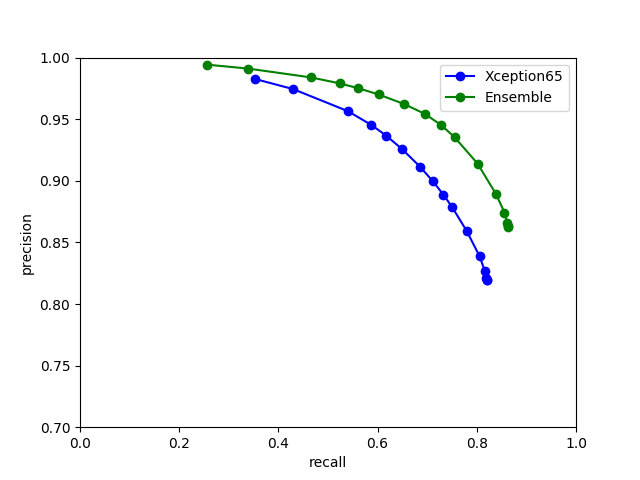}
    \caption{Precision / recall points evaluated at different confidence threshold starting from 0.1 (bottom-right points) to 0.995 (top-left points) on GTA-to-Cityscapes transfer. Note that Y-axis (precision) starts at 0.7 value to provide more detailed view.}
   \label{fig:precisionrecall}
\end{figure}

\vspace*{\floatsep}%
   

\begin{figure}[t]
\setlength{\tempwidth}{.49\linewidth}
\settoheight{\tempheight}{\includegraphics[width=\tempwidth]{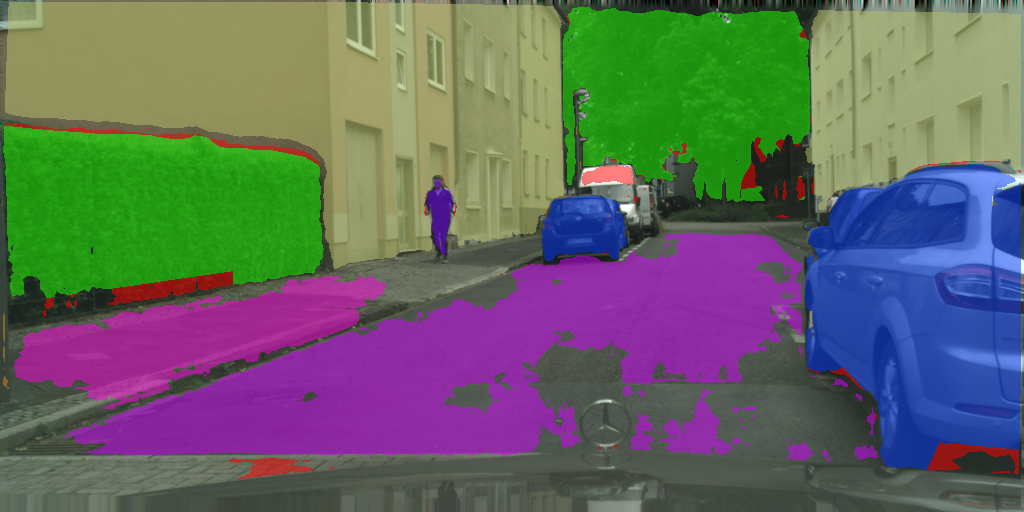}}%
\centering
\columnname{Xception65}\hfil
\columnname{Ensemble}\\
\subfloat{\includegraphics[width=\tempwidth]{imgs/fig_4_1.png}}\hfil
\subfloat{\includegraphics[width=\tempwidth]{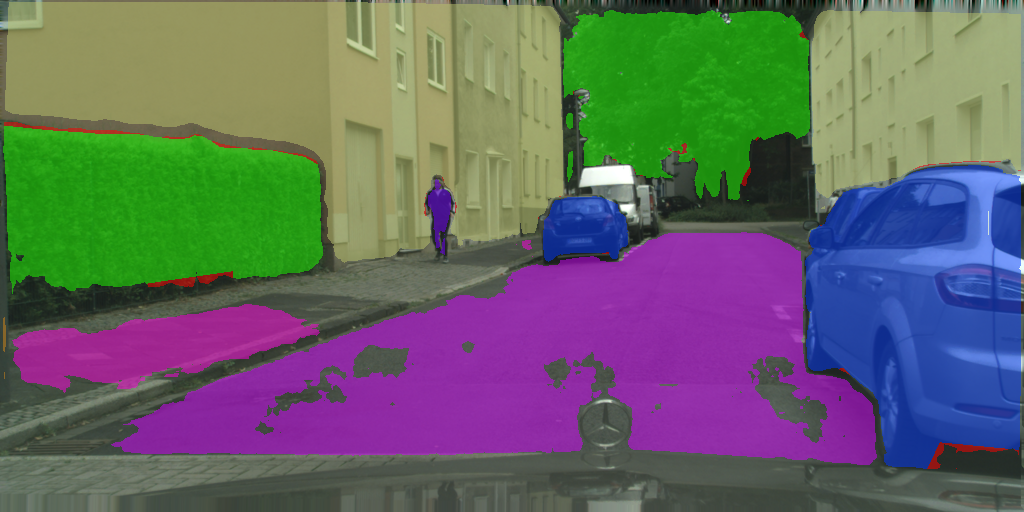}}\hfil
\subfloat{\includegraphics[width=\tempwidth]{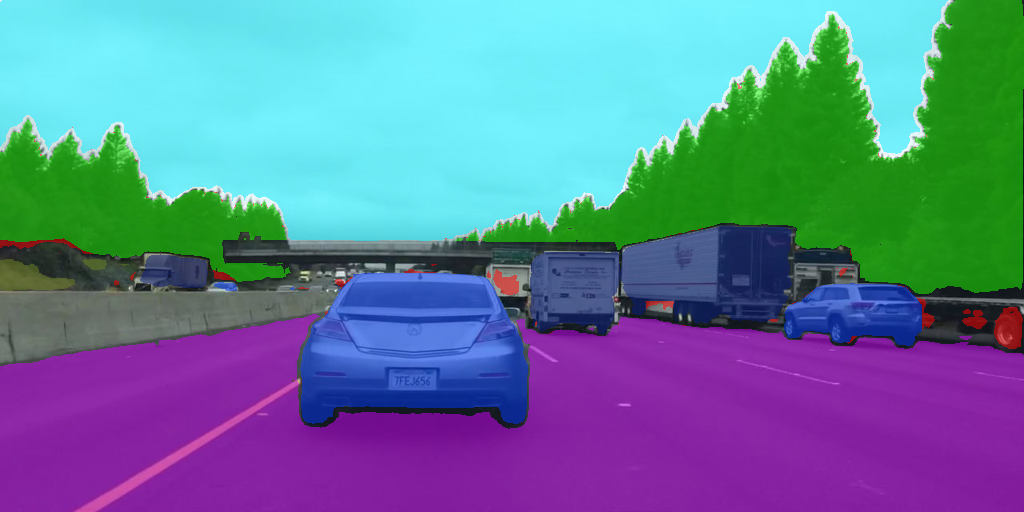}}\hfil
\subfloat{\includegraphics[width=\tempwidth]{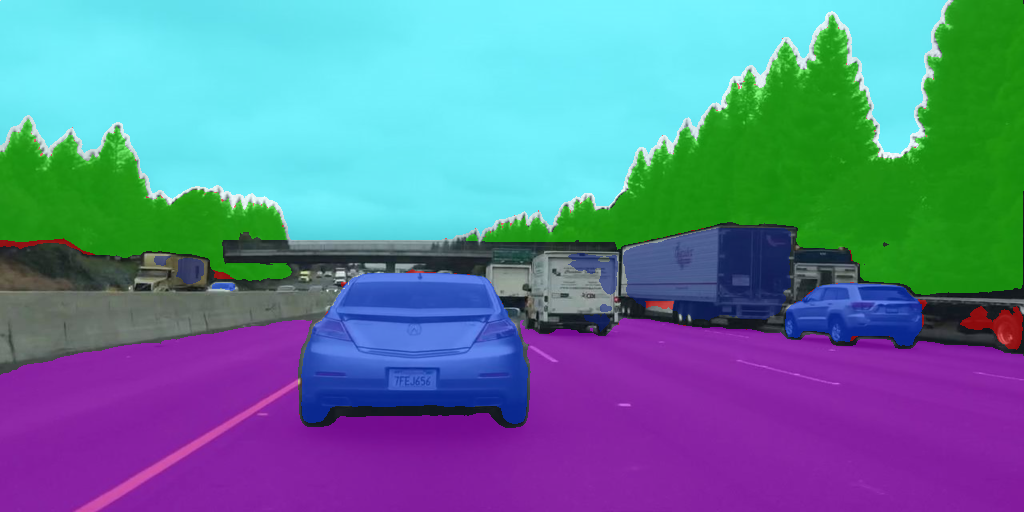}}\hfil
\caption{Examples of pseudo-labels obtained obtained on GTA-to-Cityscapes transfer (first row), and on Cityscapes-to-BDD transfer (second row). }
\label{fig:pseudo}
   
\end{figure}

One of the potential usages of well-calibrated uncertainty estimation is self-training. For this purpose, we first estimated the precision/recall points for different confidence thresholds $t$ (Fig. \ref{fig:precisionrecall}). Namely, such a curve is an approximation of how many pixels can be automatically annotated with what precision. Overall, it can be noticed that much higher recall values were obtained for the ensemble. For example, with a precision of 95\% for the Xception65 model, the recall was around 56.5\%, while it increased to 71.2\% for the ensemble. This shows that ensembles are a very powerful technique. A complementary work to ours shows that ensembles can be used to label a new dataset efficiently \cite{edf}. The ensemble was used to coarsely annotate new datasets with high accuracy, 
and then human annotators were employed to refine the initial predictions.

\newcommand\x{90}
\def\rot{\rotatebox}

\begin{table*}[ht!]
\centering
\caption{Domain adaptation results for our models with per-class evaluation.
}
\label{tab:final}
\scalebox{0.93}{
\begin{tabular}{|l|l|l|l|l|l|l|l|l|l|l|l|l|l|l|l|l|l|l|l|l|}

\hline
Name & \rot{\x}{road} & \rot{\x}{sidewalk} & \rot{\x}{building} & \rot{\x}{wall} & \rot{\x}{fence} & \rot{\x}{pole} & \rot{\x}{traffic light} & \rot{\x}{traffic sign} & \rot{\x}{vegetation} & \rot{\x}{terrain} & \rot{\x}{sky} & \rot{\x}{person} & \rot{\x}{rider} & \rot{\x}{car} & \rot{\x}{truck} & \rot{\x}{bus} & \rot{\x}{train} & \rot{\x}{motorcycle} & \rot{\x}{bicycle} & \rot{\x}{\textbf{mIoU}} \\
\hline
\multicolumn{21}{|l|}{\textbf{Gta to Cityscapes}}\\
\hline
Baseline & 29.9 & 17.4 & 62.8 & 13.2 & 14.7 & 15.5 & 26.8 & 10.7 & 79.0 & 8.4 & 47.4 & 53.5 & 10.3 & 48.2 & 25.7 & 3.04 & 0. & 11.4 & 4.5 & 25.4\\

\hline
CJ & 80.3 & 28.9 & 80.9 & 30.9 & 22.5 & 25.8 & 37.0 & 17.5 & 83.8 & 31.0 & 76.6 & \textbf{58.4} & \textbf{19.6} & 83.0 & 28.7 & 24.7 & 0. & \textbf{27.4} & \textbf{11.0} & 40.4\\
\hline
CJ + fine & 86.1 & 36.4 & 83.1 & 24.9 & 28.7 & 27.8 & \textbf{39.6} & 19.4 & 85.7 & 38.4 & 79.5 & 56.9 & 13.0 & 86.5 & 31.0 & 23.6 & 0. & 22.6 & 0.& 41.2\\
\hline
CJ + ens & \textbf{88.6} & \textbf{43.2} & \textbf{85.0} & \textbf{36.3} & \textbf{33.8} & \textbf{30.7} & 37.4 & \textbf{21.9} & \textbf{86.8} & \textbf{44.9} & \textbf{83.9} & 57.5 & 14.5 & \textbf{87.3} & \textbf{37.2} & \textbf{32.2} & 0. & 15.0 & 0. & \textbf{44.0}\\
\hline
\multicolumn{21}{|l|}{\textbf{Cityscapes to BDD}}\\
\hline
Baseline & 88.9 & 52.4 & 65.2 & 18.5 & 18.7 & 35.2 & 35.7 & 31.9 & 78.2 & 36.1 & 75.8 & 47.3 & 22.3 & 78.5 & 23.4 & 32.7 & 0. & 41.2 & 32.0 & 42.8\\
\hline
CJ & 91.8 & 54.5 & 79.9 & \textbf{19.8} & 27.1 & \textbf{41.9} & \textbf{43.3} & 43.8 & 82.5 & 39.1 & 91.4 & 58.2 & 29.7 & 85.2 & 27.7 & 25.5 & 0. & \textbf{49.1} & 42.6 & 49.1 \\
\hline
CJ + fine & 93.2 & 60.4 & \textbf{81.4} & 18.7 & 36.6 & 37.4 & 40.5 & 44.2 & 83.0 & 42.0 & 91.7 & 62.2 & 43.7 & 85.1 & 36.4 & 23.6 & 0. & 47.6 & 48.7 & 51.4 \\
\hline
CJ + ens & \textbf{94.4} & \textbf{62.5} & 81.0 & 17.5 & \textbf{37.7} & 38.6 & 38.6 & \textbf{45.5} & \textbf{85.0} & \textbf{43.2} & \textbf{92.2} & \textbf{63.2} & \textbf{46.8} & \textbf{87.1} & \textbf{42.6} & \textbf{54.7} & 0. & 44.9 & \textbf{53.4} & \textbf{54.2}\\

\hline
\end{tabular}
}
\end{table*}


\subsection{Domain adaptation}
As was shown, the ensemble of models improved the model precision in the domain adaptation setting, and greatly improved uncertainty estimation, which can be efficiently utilized in the self-training setting. First, a semantic segmentation model was used to obtain pseudo-labels on the target datasets, using some threshold $t$. In the literature, the threshold of value 0.9 is commonly used \cite{textureinvariant}, and the same value is used in our experiments. For the ensemble variant, such a threshold allows 70.1\% of the pixels to be annotated with 92.6\% accuracy, on GTA-to-Cityscapes transfer. Fig. \ref{fig:pseudo} shows obtained pseudo-labels. In general, it can be noticed that the ensemble's labels were less noisy, and the object boundaries were more tightly aligned around the object of interest.


After the pseudo-labels were obtained for the target datasets, they were used for model fine-tuning. In this section results for different models are presented:
\begin{enumerate}
    \item ResNet-101 using standard data augmentation.
    \item ResNet-101 trained using additional color jittering data augmentation. 
    \item The previous model fine-tuned on target datasets using pseudo-labels obtained by that model (\textit{CJ + fine} in the tables)
    \item ResNet-101 fine-tuned on target datasets using pseudo-labels obtained by the model ensemble (\textit{CJ + ens} in the tables)
\end{enumerate}

\begin{figure*}
\setlength{\tempwidth}{.24\linewidth}
\settoheight{\tempheight}{\includegraphics[width=\tempwidth]{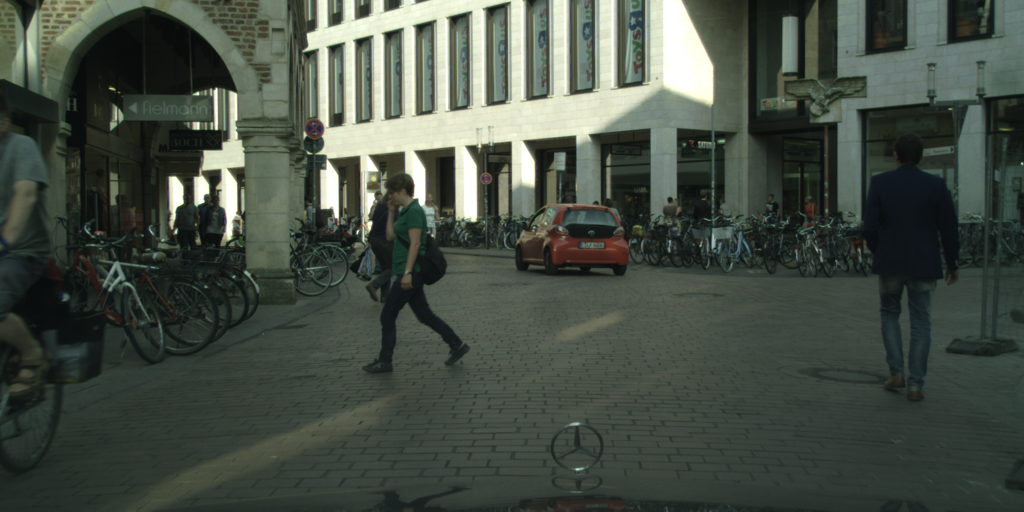}}%
\centering
\hspace{\baselineskip}
\columnname{Image}\hfil
\columnname{Finetuned}\hfil
\columnname{Ensemble finetuned}\hfil
\columnname{Ground truth}\\
\subfloat{\includegraphics[width=\tempwidth]{imgs/fig_5_1.png}}\hfil
\subfloat{\includegraphics[width=\tempwidth]{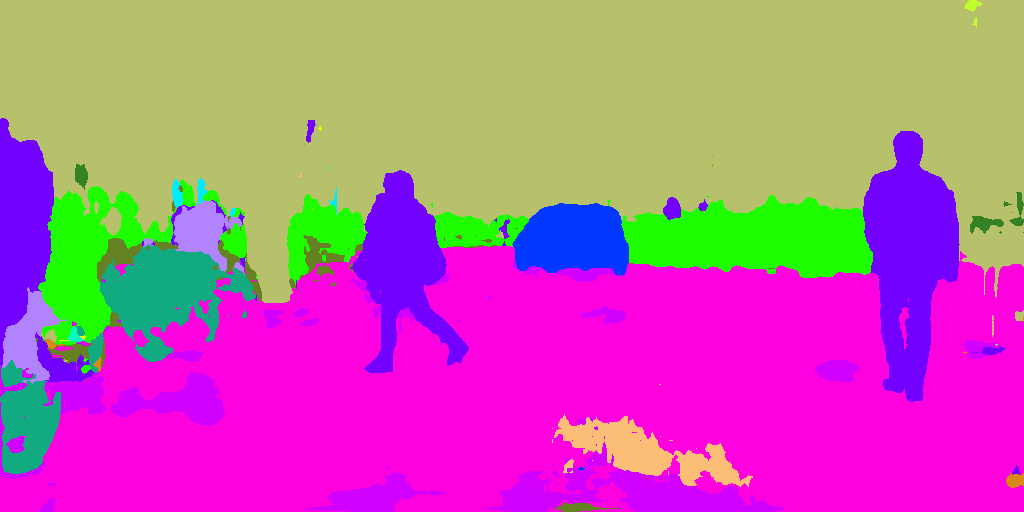}}\hfil
\subfloat{\includegraphics[width=\tempwidth]{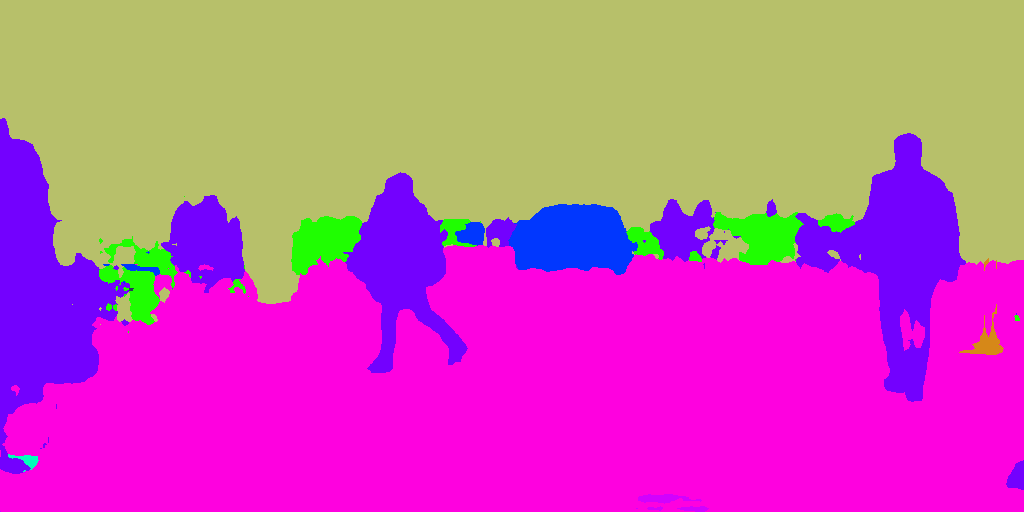}}\hfil
\subfloat{\includegraphics[width=\tempwidth]{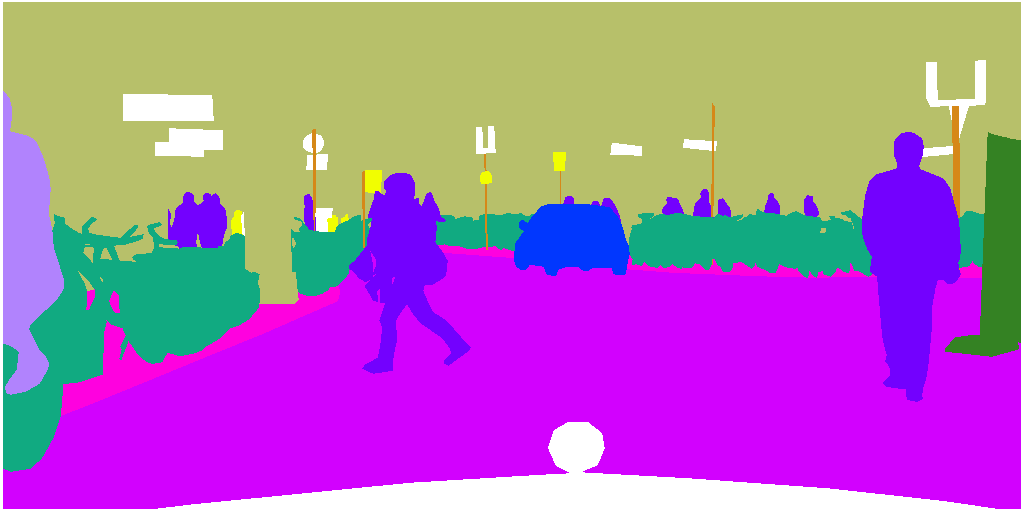}}\\
\subfloat{\includegraphics[width=\tempwidth]{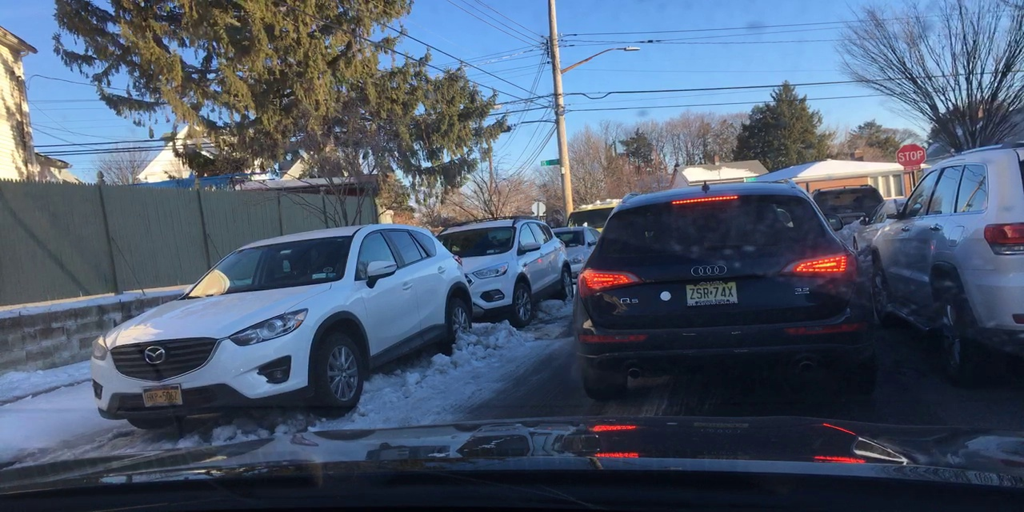}}\hfil
\subfloat{\includegraphics[width=\tempwidth]{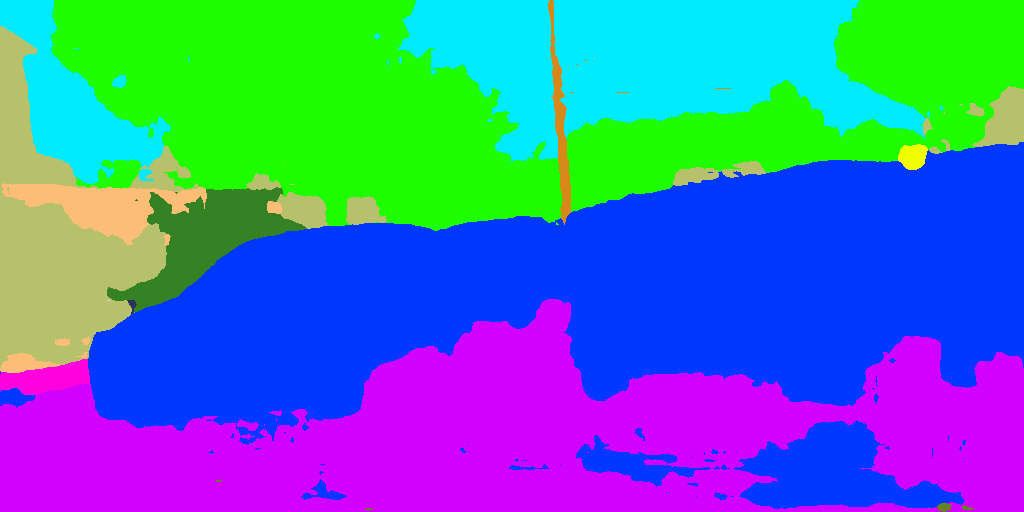}}\hfil
\subfloat{\includegraphics[width=\tempwidth]{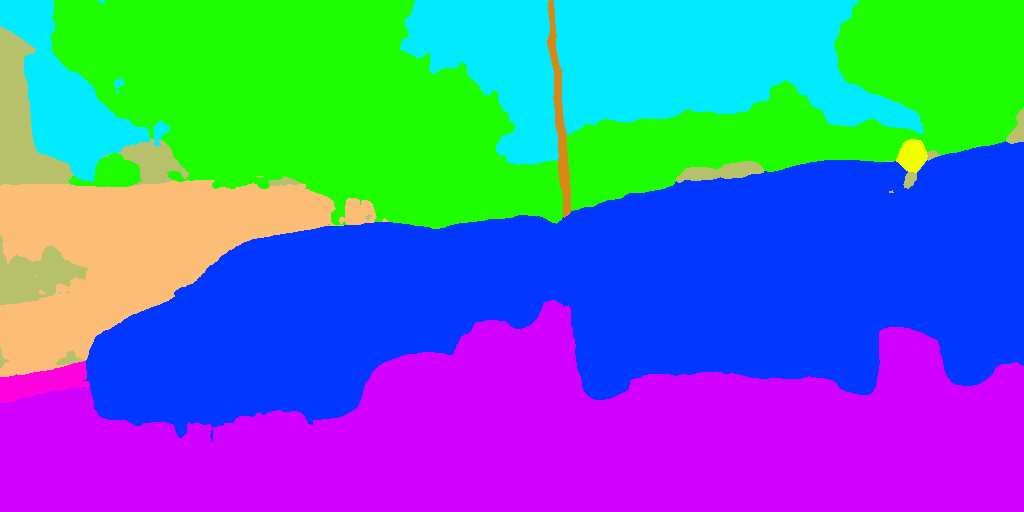}}\hfil
\subfloat{\includegraphics[width=\tempwidth]{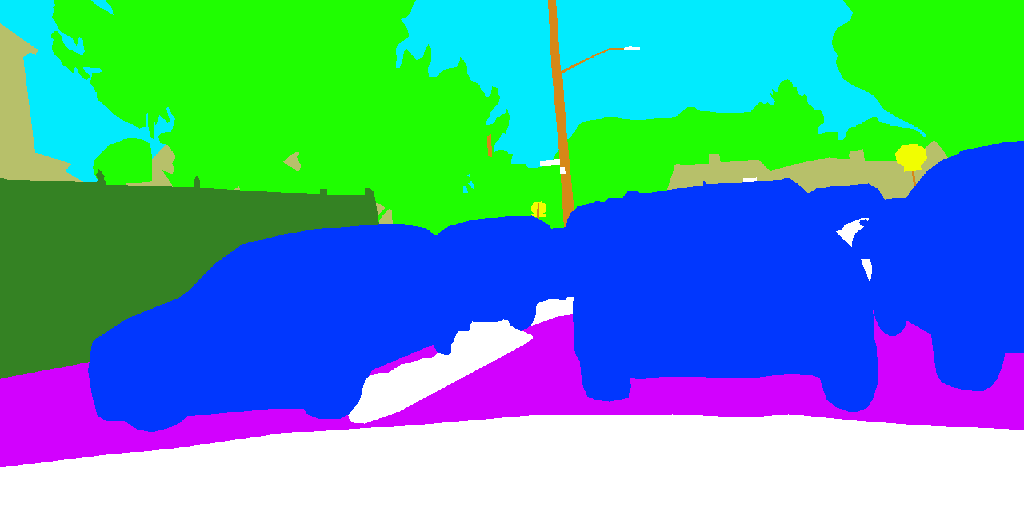}} \\

\subfloat{\includegraphics[width=\tempwidth]{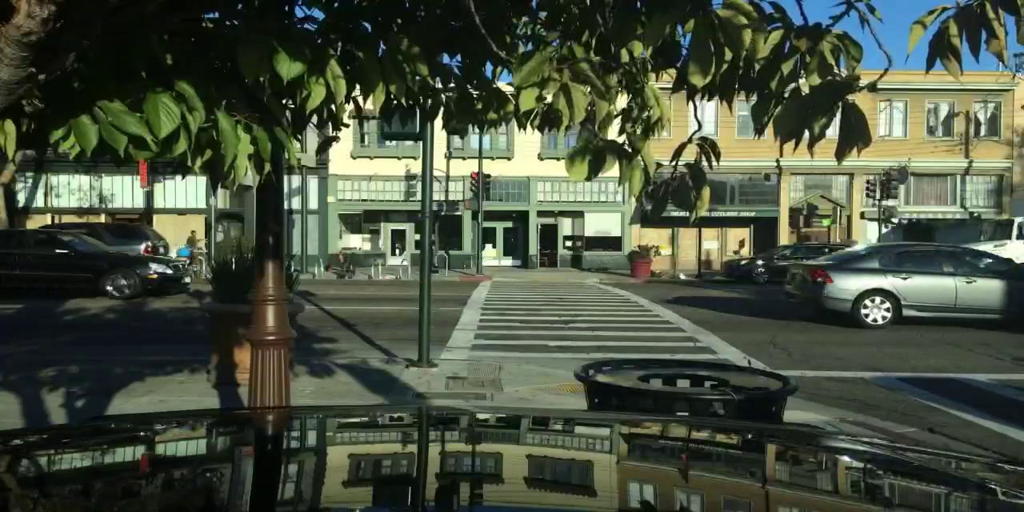}}\hfil
\subfloat{\includegraphics[width=\tempwidth]{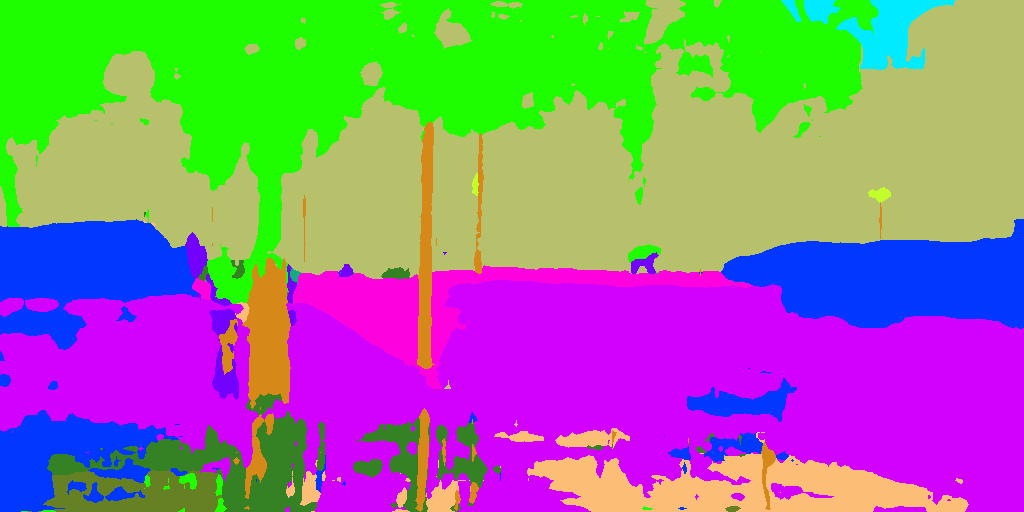}}\hfil
\subfloat{\includegraphics[width=\tempwidth]{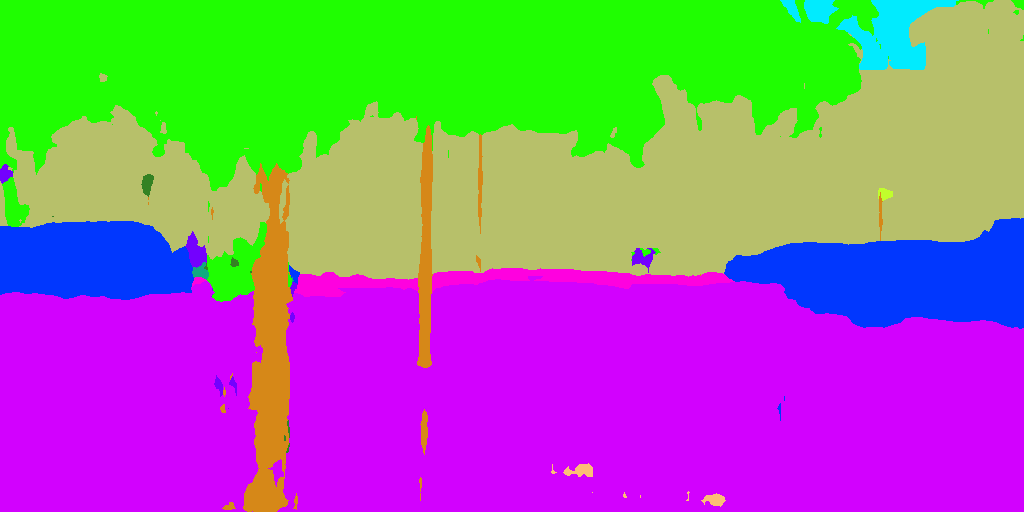}}\hfil
\subfloat{\includegraphics[width=\tempwidth]{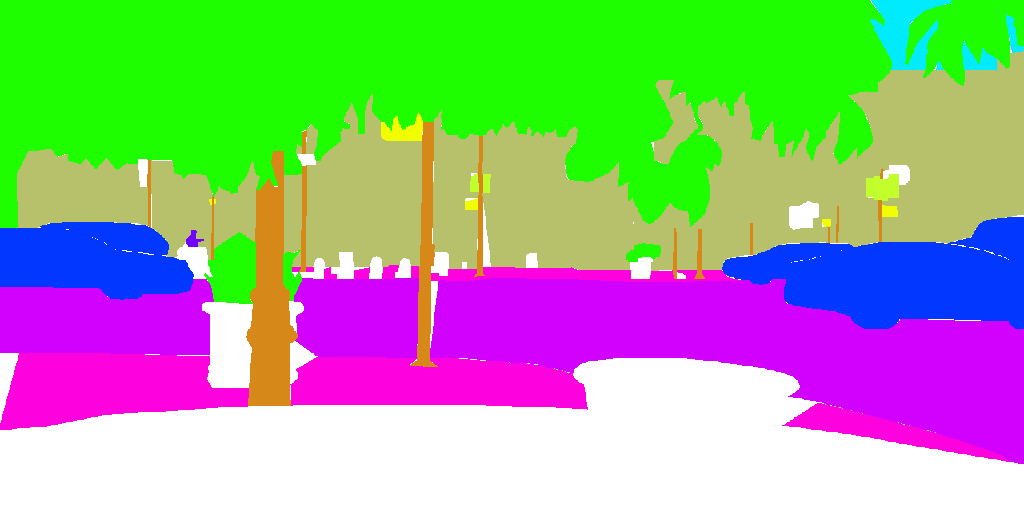}} 

\caption{Qualitative results of trained models on GTA-to-Cityscapes transfer (first row) and Cityscapes-to-BDD transfer (consecutive rows). White color corresponds to the ignore label.}
\label{fig:results}
\end{figure*}

Table \ref{tab:final} shows the final results, including per-class evaluation. Firstly, consistent with other works, the self-training approach improved the model accuracy (from 40.4 to 41.2, and from 49.1 to 51.4 on the Cityscapes and BDD datasets, respectively). When the pseudo-labels were collected using an ensemble approach, the model accuracy was further greatly increased (from 41.2 to 44.0 on Cityscapes dataset, and from 51.4 to 54.2 on BDD dataset). Fig. \ref{fig:results} shows qualitative results. In general it can be noticed that obtained segmentation maps are less noisy, especially in more challenging cases (second and third row).

As a sanity check, the fine-tuning was also performed on the highest-capacity model (Xception65). In that case the mIoU has increased from 41.3 to 45.3, and from 52.4 to 53.6 on Cityscapes and BDD datasets, respectively. This shows that an ensemble approach is effective, also when the finetuned model is a very strong member of the ensemble.

In general our results are very promising. Ensemble approach turned out to be very effective in terms out model accuracy and uncertainty estimation, even for the large distributional shift (GTA to Cityscapes). On the drawback side, ensembling did not improve for classes with the lowest precision (bicycle, motorcycle, rider). It might occur because the baseline model is a week detector of such classes, so as a result, there are very few pseudo-labels collected for those classes. Improving accuracy for such classes remains an open challenge in self-training methods.
Another important problem with an ensemble is that multiple models have to be trained and evaluated, which is very costly. However, the recently introduced BatchEnsemble method significantly reduced the computational and memory costs \cite{batchensemble}. Similarly, it was shown that training one neural network with a multi-input multi-output (MIMO) configuration could be an efficient strategy to improve the models’ robustness \cite{mimo}. However, applying those ideas to the high-level task of semantic segmentation is important future work.




\section{Conclusions}

In this work, calibration of model predictive uncertainty under different, realistic real-world application settings was studied. It was shown that the ensemble of models significantly improved the uncertainty estimation and overall accuracy, especially under domain shift. Notably, the performance gains are consistent even when the domain gap is large (simulation-to-real transfer case). Our ensemble consists of models using different backbones and/or data augmentations. Interestingly, it was also shown that simple color transformations can lead to a similar performance improvement as much more sophisticated style-transfer augmentation, and that both types of data augmentation are crucial in the domain adaptation setting, which confirms and extends recent findings \cite{rethinking}.

Further, the ensemble of models was utilized for domain adaptation using the self-training method. The improved uncertainty calibration and model accuracy allowed the fine tuning stage to be improved significantly, since the mIoU increased from 41.2 to 44.0, and from 51.4 to 54.2 on the Cityscapes and BDD datasets, respectively.  Our approach is complementary to other domain adaptation methods based on the self-training, thus it could be easily combined with them, providing an interesting future work subject.

\bibliography{biblio.bib}

\begin{thebibliography}{10}

\bibitem{concrete}
D.~Amodei, C.~Olah, J.~Steinhardt, P.~Christiano, J.~Schulman, and D.~Man{\'e},
  ``Concrete problems in ai safety,'' {\em arXiv preprint arXiv:1606.06565},
  2016.

\bibitem{dodge}
S.~Dodge and L.~Karam, ``A study and comparison of human and deep learning
  recognition performance under visual distortions,'' in {\em 26th
  international conference on computer communication and networks (ICCCN)},
  pp.~1--7, IEEE, 2017.

\bibitem{objectnet}
A.~Barbu {\em et~al.}, ``Objectnet: A large-scale bias-controlled dataset for
  pushing the limits of object recognition models,'' in {\em Advances in Neural
  Information Processing Systems}, pp.~9453--9463, 2019.

\bibitem{temporal}
S.~H. {Abdulhussain}, S.~A.~R. {Al-Haddad}, M.~I. {Saripan}, B.~M. {Mahmmod},
  and A.~{Hussien}, ``Fast temporal video segmentation based on
  krawtchouk-tchebichef moments,'' {\em IEEE Access}, vol.~8, pp.~72347--72359,
  2020.

\bibitem{winter}
C.~Michaelis, B.~Mitzkus, R.~Geirhos, E.~Rusak, O.~Bringmann, A.~S. Ecker,
  M.~Bethge, and W.~Brendel, ``Benchmarking robustness in object detection:
  Autonomous driving when winter is coming,'' {\em CoRR}, vol.~abs/1907.07484,
  2019.

\bibitem{cnnbiased}
R.~Geirhos {\em et~al.}, ``Imagenet-trained cnns are biased towards texture;
  increasing shape bias improves accuracy and robustness,'' in {\em 7th
  International Conference on Learning Representations, ICLR}, 2019.

\bibitem{HendrycksD19}
D.~Hendrycks and T.~G. Dietterich, ``Benchmarking neural network robustness to
  common corruptions and perturbations,'' in {\em 7th International Conference
  on Learning Representations, {ICLR}}, 2019.

\bibitem{calibration}
C.~Guo, G.~Pleiss, Y.~Sun, and K.~Q. Weinberger, ``On calibration of modern
  neural networks,'' in {\em Proceedings of the 34th International Conference
  on Machine Learning-Volume 70}, pp.~1321--1330, 2017.

\bibitem{trustuncertainty}
Y.~Ovadia {\em et~al.}, ``Can you trust your model's uncertainty? evaluating
  predictive uncertainty under dataset shift,'' in {\em Advances in Neural
  Information Processing Systems}, pp.~13991--14002, 2019.

\bibitem{multi}
D.~{Feng}, Y.~{Cao}, L.~{Rosenbaum}, F.~{Timm}, and K.~{Dietmayer},
  ``Leveraging uncertainties for deep multi-modal object detection in
  autonomous driving,'' in {\em 2020 IEEE Intelligent Vehicles Symposium (IV)},
  pp.~877--884, 2020.

\bibitem{noisystudent}
Q.~Xie, M.~Luong, E.~H. Hovy, and Q.~V. Le, ``Self-training with noisy student
  improves imagenet classification,'' in {\em {IEEE/CVF} Conference on Computer
  Vision and Pattern Recognition,}, pp.~10684--10695, 2020.

\bibitem{ensemblesuncertainty}
F.~K. Gustafsson, M.~Danelljan, and T.~B. Sch{\"o}n, ``Evaluating scalable
  bayesian deep learning methods for robust computer vision,'' in {\em
  Proceedings of the IEEE/CVF Conference on Computer Vision and Pattern
  Recognition (CVPR) Workshops}, 2020.

\bibitem{textureinvariant}
M.~Kim and H.~Byun, ``Learning texture invariant representation for domain
  adaptation of semantic segmentation,'' in {\em Proceedings of the IEEE/CVF
  Conference on Computer Vision and Pattern Recognition}, pp.~12975--12984,
  2020.

\bibitem{selfeccv2018}
Y.~Zou, Z.~Yu, B.~V. K.~V. Kumar, and J.~Wang, ``Unsupervised domain adaptation
  for semantic segmentation via class-balanced self-training,'' in {\em The
  European Conference on Computer Vision (ECCV)}, vol.~11207, pp.~297--313,
  Springer, 2018.

\bibitem{semanticbenchmark}
C.~Kamann and C.~Rother, ``Benchmarking the robustness of semantic segmentation
  models,'' in {\em Proceedings of the IEEE/CVF Conference on Computer Vision
  and Pattern Recognition}, pp.~8828--8838, 2020.

\bibitem{Cygert}
S.~{Cygert} and A.~{Czyżewski}, ``Toward robust pedestrian detection with data
  augmentation,'' {\em IEEE Access}, vol.~8, pp.~136674--136683, 2020.

\bibitem{todorealistic}
A.~Oliver, A.~Odena, C.~Raffel, E.~D. Cubuk, and I.~J. Goodfellow, ``Realistic
  evaluation of deep semi-supervised learning algorithms,'' in {\em Advances in
  Neural Information Processing Systems}, pp.~3235–--3246, 2018.

\bibitem{robcompr}
S.~Cygert and A.~Czyzewski, ``Robustness in compressed neural networks for
  object detection,'' {\em CoRR}, vol.~abs/2102.05509, 2021.

\bibitem{temperature}
J.~Platt {\em et~al.}, ``Probabilistic outputs for support vector machines and
  comparisons to regularized likelihood methods,'' {\em Advances in large
  margin classifiers}, vol.~10, no.~3, pp.~61--74, 1999.

\bibitem{Gal}
Y.~Gal and Z.~Ghahramani, ``Dropout as a bayesian approximation: Representing
  model uncertainty in deep learning,'' in {\em Proceedings of the 33nd
  International Conference on Machine Learning, ICML}, vol.~48, pp.~1050--1059,
  2016.

\bibitem{ensemble}
B.~Lakshminarayanan, A.~Pritzel, and C.~Blundell, ``Simple and scalable
  predictive uncertainty estimation using deep ensembles,'' in {\em Advances in
  Neural Information Processing Systems 30 (NeurIPS)}, pp.~6402--6413, 2017.

\bibitem{ensembles_1990}
L.~K. {Hansen} and P.~{Salamon}, ``Neural network ensembles,'' {\em IEEE
  Transactions on Pattern Analysis and Machine Intelligence}, vol.~12, no.~10,
  pp.~993--1001, 1990.

\bibitem{TODOnipstransfer}
B.~Neyshabur, H.~Sedghi, and C.~Zhang, ``What is being transferred in transfer
  learning?,'' in {\em Advances in Neural Information Processing Systems 30
  (NeurIPS)}, 2020.

\bibitem{histogram}
A.~Abramov, C.~Bayer, and C.~Heller, ``Keep it simple: Image statistics
  matching for domain adaptation,'' {\em CoRR}, vol.~abs/2005.12551, 2020.

\bibitem{james2019simtoreal}
S.~James, P.~Wohlhart, M.~Kalakrishnan, D.~Kalashnikov, A.~Irpan, J.~Ibarz,
  S.~Levine, R.~Hadsell, and K.~Bousmalis, ``Sim-to-real via sim-to-sim:
  Data-efficient robotic grasping via randomized-to-canonical adaptation
  networks,'' in {\em IEEE Conference on Computer Vision and Pattern
  Recognition}, pp.~12627--–12637, 2019.

\bibitem{synthetic}
H.~{Zhang}, Y.~{Tian}, K.~{Wang}, H.~{He}, and F.~{Wang}, ``Synthetic-to-real
  domain adaptation for object instance segmentation,'' in {\em 2019
  International Joint Conference on Neural Networks (IJCNN)}, pp.~1--7, 2019.

\bibitem{medicalcaluncertainty}
A.~{Mehrtash}, W.~M. {Wells}, C.~M. {Tempany}, P.~{Abolmaesumi}, and
  T.~{Kapur}, ``Confidence calibration and predictive uncertainty estimation
  for deep medical image segmentation,'' {\em IEEE Transactions on Medical
  Imaging}, vol.~39, no.~12, pp.~3868--3878, 2020.

\bibitem{modelcompression}
C.~Bucila, R.~Caruana, and A.~Niculescu{-}Mizil, ``Model compression,'' in {\em
  Proceedings of the Twelfth {ACM} {SIGKDD} International Conference on
  Knowledge Discovery and Data Mining}, pp.~535--541, {ACM}, 2006.

\bibitem{FCN}
E.~Shelhamer, J.~Long, and T.~Darrell, ``Fully convolutional networks for
  semantic segmentation,'' {\em {IEEE} Trans. Pattern Anal. Mach. Intell.},
  vol.~39, no.~4, pp.~640--651, 2017.

\bibitem{deeplab}
L.-C. Chen, Y.~Zhu, G.~Papandreou, F.~Schroff, and H.~Adam, ``Encoder-decoder
  with atrous separable convolution for semantic image segmentation,'' in {\em
  The European Conference on Computer Vision (ECCV)}, 2018.

\bibitem{style_transfer}
X.~Huang and S.~Belongie, ``Arbitrary style transfer in real-time with adaptive
  instance normalization,'' in {\em Proceedings of the IEEE International
  Conference on Computer Vision}, pp.~1501--1510, 2017.

\bibitem{playingfordata}
S.~R. Richter, V.~Vineet, S.~Roth, and V.~Koltun, ``Playing for data: Ground
  truth from computer games,'' in {\em European conference on computer vision},
  pp.~102--118, Springer, 2016.

\bibitem{cityscapes}
M.~Cordts {\em et~al.}, ``The cityscapes dataset for semantic urban scene
  understanding,'' in {\em Proc. of the IEEE Conference on Computer Vision and
  Pattern Recognition (CVPR)}, 2016.

\bibitem{bdd}
F.~Yu {\em et~al.}, ``{BDD100K:} {A} diverse driving dataset for heterogeneous
  multitask learning,'' in {\em 2020 {IEEE/CVF} Conference on Computer Vision
  and Pattern Recognition}, pp.~2633--2642, {IEEE}, 2020.

\bibitem{origins}
K.~L. Hermann, T.~Chen, and S.~Kornblith, ``The origins and prevalence of
  texture bias in convolutional neural networks,'' in {\em Advances in Neural
  Information Processing Systems 33, NeurIPS}, 2020.

\bibitem{pnas}
C.~Liu, B.~Zoph, M.~Neumann, J.~Shlens, W.~Hua, L.~Li, L.~Fei{-}Fei, A.~L.
  Yuille, J.~Huang, and K.~Murphy, ``Progressive neural architecture search,''
  in {\em Computer Vision - {ECCV} 2018 - 15th European Conference},
  pp.~19--35, 2018.

\bibitem{edf}
G.~Zhou, S.~Dulloor, D.~G. Andersen, and M.~Kaminsky, ``{EDF:} ensemble,
  distill, and fuse for easy video labeling,'' {\em CoRR}, vol.~abs/1812.03626,
  2018.

\bibitem{batchensemble}
Y.~Wen, D.~Tran, and J.~Ba, ``Batchensemble: an alternative approach to
  efficient ensemble and lifelong learning,'' in {\em 8th International
  Conference on Learning Representations, {ICLR}}, 2020.

\bibitem{mimo}
M.~Havasi {\em et~al.}, ``Training independent subnetworks for robust
  prediction,'' in {\em International Conference on Learning Representations,
  ICLR}, 2021.

\bibitem{rethinking}
F.~C. Borlino, A.~D'Innocente, and T.~Tommasi, ``Rethinking domain
  generalization baselines,'' in {\em 25th International Conference on Pattern
  Recognition, {ICPR}}, 2020.

\end{thebibliography}
\bibliographystyle{ieeetr}

\end{document}